\documentclass{llncs}

\usepackage{algorithm}
\usepackage{algorithmic}
\usepackage{amsmath}
\usepackage{caption}
\usepackage{graphicx}
\usepackage{wrapfig}

\begin{document}

\title{Generating Accurate Virtual Examples For Lifelong Machine Learning}

\author{Sazia Mahfuz 
}

\institute{Queen's University, \email{sazia.mahfuz@queensu.ca}
}

\maketitle              

\begin{abstract}
Lifelong machine learning (LML) is an area of machine learning research concerned with human-like persistent and cumulative nature of learning. LML system's objective is consolidating new information into an existing machine learning model without catastrophically disrupting the prior information. Our research addresses this LML retention problem for creating a knowledge consolidation network through task rehearsal without retaining the prior task's training examples. We discovered that the training data reconstruction error from a trained Restricted Boltzmann Machine can be successfully used to generate accurate virtual examples from the reconstructed set of a uniform random set of examples given to the trained model. We also defined a measure for comparing the probability distributions of two datasets given to a trained network model based on their reconstruction mean square errors. 
\end{abstract}

\section{Introduction and Background} \label{Introduction}
Humans learn new knowledge as they grow older while retaining prior knowledge. Similarly, LML systems can retain prior tasks' knowledge over a long time while integrating, or consolidating new task's knowledge periodically \cite{Silver-Yang-2013}. Just like living beings, \textit{consolidation} enables the integration of new information into the existing learning system. The main challenge in consolidation is \textit{catastrophic forgetting} and overcoming the \textit{stability-plasticity dilemma}. 
\par
The process of retaining the old information and yet being able to integrate the new information is known as the \textit{stability-plasticity dilemma}. A neural network learning mechanism capable of stability and plasticity can rehearse examples of prior knowledge to maintain functional stability while slowly changing its representation to accommodate new knowledge \cite{Silver-Mason-Ejabu-2015}. 
\par
\textit{Catastrophic forgetting} can be defined as the disruption or loss of the prior training information while integrating new information to a trained model \cite{Robins-1995}. The challenge is to reduce the affect of catastrophic forgetting, which can be done by the rehearsal of a subset of the old information forcing the learning system to retain the structure of the old information \cite{Robins-1995}. One approach is to store a set of examples from the training set for each task. But the space complexity of this approach will grow linearly with the number of tasks, each time lengthening the training time for a new task. An alternate approach is to use the concept of sweep pseudorehearsal discussed by Robins \cite{Robins-1996}; where examples are created by passing randomly created input vector through the learning system, the generated outputs are recorded for that particular input vector, and these are randomly included in the training session of the new task. From these reconstructed examples if we select only those examples which adhere to the probability distribution of the training data, then those selected examples are referred to as virtual examples (VEs) \cite{Silver-Et-Al-2008} in our research.
\par
The focus of this research is to develop a method by which to generate VEs of prior tasks from an existing neural network model such that the example(s) adhere to the probability distribution of those prior tasks. Our approach was to investigate feasible approaches for generating accurate VEs, and then evaluate the selected methods based on the VEs' based on their adherence to the prior task distribution. The research aimed to generate VEs adhering to the input variables' distributions.

\section{Related Work} \label{Related}
Research by Kirkpatrick et al. \cite{Kirkpatrick-17} presented a novel algorithm, elastic weight consolidation (EWC) by decreasing the weight plasticity which avoided the catastrophic forgetting of old training information during the integration of new information. He et al \cite{Xu-2018} proposed a variant of the backpropagation algorithm, ``conceptor-aided backprop'', where conceptors were used to protect the gradients from degradation of prior trained tasks. Compared with the above approaches, our method used the Robins' pseudorehearsal approach to handle catastrophic forgetting. 

\section{Generating VEs Based on a trained Restricted Boltzmann Machine (RBM) Reconstruction Error}	 \label{Theory}	
An accurately trained  Restricted Boltzmann Machine (RBM) reconstructs the training data with a low error. This observation led us to investigate that after one oscillation, which is just passing the data to the trained model and recording the reconstruction, of feeding a uniform random set of examples into the model, the examples that are closer to the training data distribution are more accurately reconstructed than the other examples. This finding suggested that we can select the VEs based on the reconstruction error. 
\par
Because all of the generated examples from the uniform random set of examples do not adhere to the training data distribution after one oscillation, an approach was taken to select the examples such that a tolerance level was satisfied. Mean Squared Error (MSE) for the training data reconstruction was used as the initial tolerance level. If the sum of squared error for the uniform random input example and its corresponding reconstruction fell below the defined tolerance value, then the example was considered a VE.
\par
\textbf{Selecting the tolerance level}	\label{Tolerance}
For one, two, four-dimensional input data, the tolerance level had been selected using the MSE between the training data and its corresponding reconstruction.  
\par
\textbf{Success Criterion}	\label{SuccessCriterion}
To evaluate the success of the adherence of the virtual example distribution to the training data distribution, that is to evaluate the accuracy of the VEs, we defined a measure using the reconstruction error from a trained model, called the \textit{Autoencoder-based Divergence Measure}. 
We considered unsupervised autoencoder approach to measure the difference between two probability distributions. The idea was that if our model had been accurately trained, we could measure the relative degree of similarity between an example set and the original training set's probability distributions. We defined the measure called the Autoencoder-based Divergence Measure (ADM) as follows:
Let, $MSE_{TRN}$ = the MSE of the training data on the RBM model and $MSE_{TST}$ = the MSE of the test data given to the trained RBM model, then
\begin{equation}
ADM = \frac{MSE_{TST}}{MSE_{TRN}}
\end{equation}
We verified that, ``$0 < ADM <=1$'' signifies same probability distribution as the training data; \newline ``$1< ADM <2$'' signifies similar or partial space of the training data; \newline ``$ADM \geq 2$'' signifies increasingly different probability distribution than the training data.
\par
\textbf{Experiment - Four-dimensional Data:} 
The four-dimensional synthetic input data had 1000 examples. The input data was selected in such a way that there are various numbers of distinct regions in each of the dimensions within the range of 0 and 1. 
\par
For this experiment, the tolerance level measured by MSE was 0.000395, which was the MSE between the training data and its corresponding reconstruction after one oscillation. Thus 80 virtual examples were selected from the reconstructed set of 5000 uniform random examples passed to the model trained on the training examples $x_{1(1T)}$, $x_{2(1T)}$, $x_{3(1T)}$, $x_{4(1T)}$.
\par
\begin{wrapfigure}{R}{0.4\textwidth}
	\begin{center}
		\includegraphics[width=6cm,keepaspectratio]{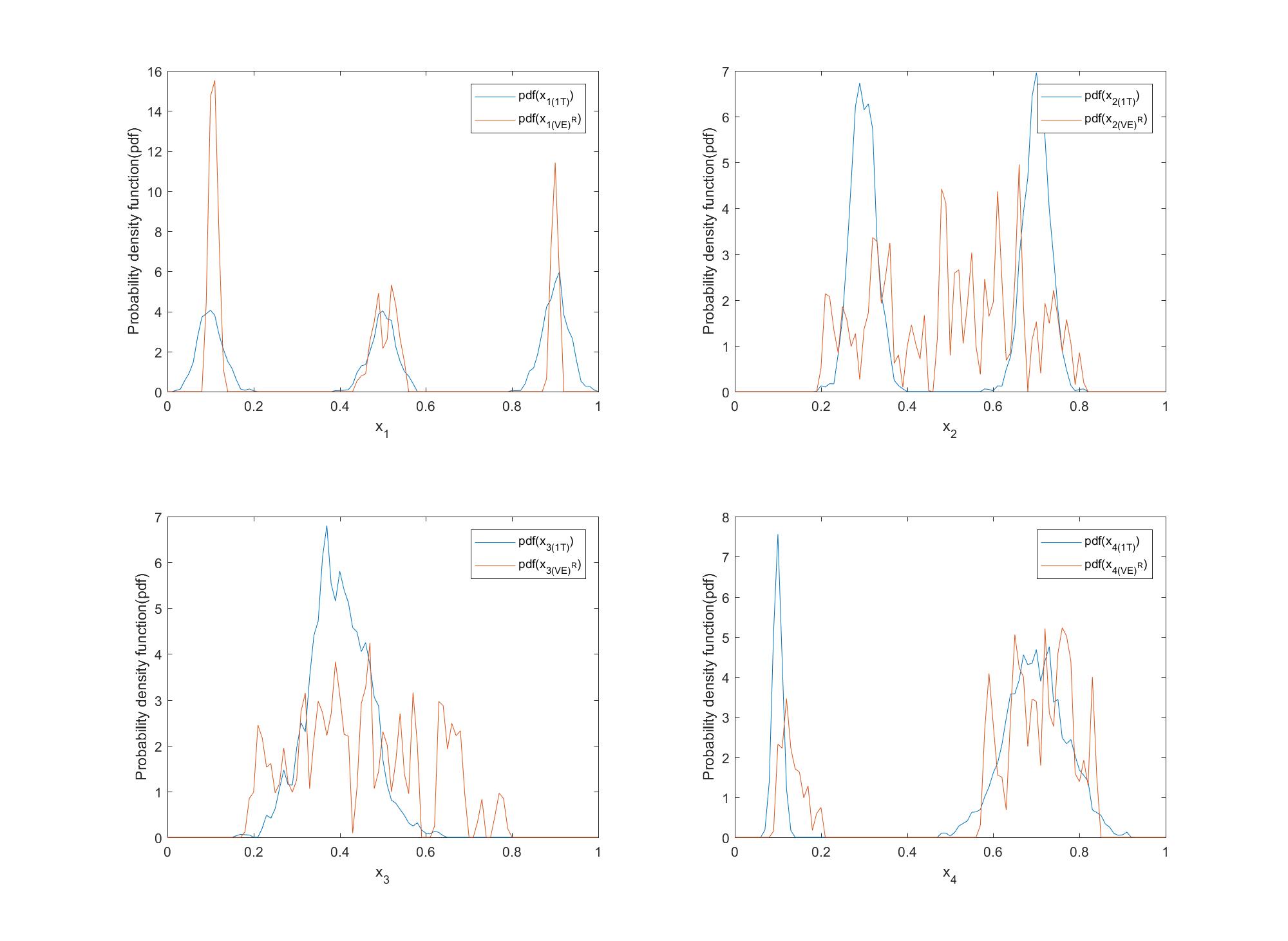}
		\captionof{figure}{The probability density funcitons for the training examples and the selected virtual examples}
		\label{fig:Exp1_4D_VE_MSE_hist}
	\end{center}
\end{wrapfigure}

We calculated the value of ADM as 
ADM = $\frac{MSE_{VE}}{MSE_{TRN}}$ = $\frac{0.000228}{0.000395}$ = 0.5772.
This value signified that the virtual examples are from the same probability distribution as the training data.
In Figure \ref{fig:Exp1_4D_VE_MSE_hist}, the blue probability density functions represent the training data. The red probability density functions represent the virtual examples selected from the reconstructed set of 5000 uniform random set of examples after one oscillation using the trained model.
\par
\textbf{Discussion: }	
The following discusses the merits of the measure based on the experimental results.
\par
For one, two, four-dimensional data, tolerance level measured by MSE had been successful in generating virtual examples validated by the results of the autoencoder-based divergence measure. The success criterion resulted that the selected examples adhere to the training data distribution. So we can consider these examples as VEs for our purpose, and we can use this measure for generating accurate VEs. 

\section{Conclusion} \label{Conclusion}
The important findings of this research are as follows:
\newline
1. We investigated approaches for generating VEs from the reconstruction of a uniform random set of examples passed through a generative RBM model. Empirically, we found that we can generate accurate VEs validated by autoencoder-based divergence measure from uniform random examples based on the reconstruction error measured by MSE.  
\newline
2. We defined and successfully verified a metric called autoencoder-based divergence measure for comparing the probability distributions of two given datasets using the reconstruction MSE from the trained RBM model. 

\section*{Acknowledgement}
I would like to express my sincerest gratitude towards my supervisor Dr. Daniel L. Silver for his consistent, constant support and patience throughout my graduate studies and his valuable comments on writing this summary.

\bibliographystyle{splncs}      
\bibliography{cai_2019}            

\end{document}